\documentclass[conf]{new-aiaa}
\usepackage[utf8]{inputenc}

\usepackage{graphicx}
\usepackage{amsmath}
\usepackage[version=4]{mhchem}
\usepackage{siunitx}
\usepackage{longtable,tabularx}
\usepackage{float}

\title{Dynamic Scene 3D Reconstruction of an Uncooperative Resident Space Object}

\author{Bala Prenith Reddy Gopu,\footnote{PhD Student, Department of Aerospace, Physics and Space Sciences, bgopu2023@my.fit.edu}, Timothy Jacob Huber,\footnote{PhD Student, Department of Aerospace, Physics and Space Sciences, thuber2019@my.fit.edu}, George M. Nehma,\footnote{PhD Student, Department of Aerospace, Physics and Space Sciences, gnehma2020@my.fit.edu}, Patrick Quinn,\footnote{PhD Student, Department of Aerospace, Physics and Space Sciences, pquinn2019@my.fit.edu} and Madhur Tiwari\footnote{Assistant Professor, Department of Aerospace, Physics and Space Sciences, mtiwari@fit.edu}}
\affil{Florida Institute of Technology, Melbourne, Florida, 32901}

\author{Matt Ueckermann,\footnote{Engineer, Creare LLC}, David Hinckley,\footnote{Engineer, Creare LLC}, Christopher McKenna,\footnote{Engineer, Creare LLC}}
\affil{Creare LLC, Hanover, New Hampshire, 03755}
\begin{document}

\maketitle

\begin{abstract}


 Characterization of uncooperative Resident Space Objects (RSO) play a crucial role in On-Orbit Servicing (OOS) and Active Debris Removal (ADR) missions to assess the geometry and motion properties. To address the challenges of reconstructing tumbling uncooperative targets, this study evaluates the performance of existing state-of-the-art 3D reconstruction algorithms for dynamic scenes, focusing on their ability to generate geometrically accurate models with high-fidelity. To support our evaluation, we developed a simulation environment using Isaac Sim to generate physics-accurate 2D image sequences of tumbling satellite under realistic orbital lighting conditions. Our preliminary results on static scenes using Neuralangelo demonstrate promising reconstruction quality. The generated 3D meshes closely match the original CAD models with minimal errors and artifacts when compared using Cloud Compare (CC). The reconstructed models were able to capture critical fine details for mission planning. This provides a baseline for our ongoing evaluation of dynamic scene reconstruction.
 
\end{abstract}

\section{Nomenclature}
{\renewcommand\arraystretch{1.0}
\noindent\begin{longtable*}{@{}l @{\quad=\quad} l@{}}
$RSO$  & Resident Space Object \\
$RPO$  & Rendezvous and Proximity Operations \\
$ADR$  & Active Debris Removal \\
$OOS$  & On-orbit Servicing \\
$CW$  & Clohessy–Wiltshire \\
$NeRF$  & Neural Radiance Fields \\
$3DGS$  & 3D Gaussian Splatting \\
$NVS$  & Novel View Synthesis \\
$D-NeRF$  & Dynamic NeRF \\
$SfM$   & Structure from Motion \\
$PSNR$  & Peak Signal-to-Noise Ratio \\
$SSIM$   & Structural Similarity Index Measure \\
$CC$  & Cloud Compare \\
$CNN$  & Convolutional Neural Network \\
$DESDynl$  & Deformation, Ecosytem Structure and Dynamics of Ice \\
$GOES-R$  & Geostationary Operational Environmental Satellites \\
\end{longtable*}}

\pagebreak
\section{Introduction}

With the exponential growth of space launches every year, the number of space objects also increases, particularly in Low Earth Orbit \cite{esa2024}. This proliferation has created an increasingly complex and crowded orbital environment alongside operational satellites. Active satellites have to perform frequent maneuvers to avoid collisions with uncooperative space objects to prevent additional debris that would further congest the environment \cite{liou2011}. To address these escalating challenges, effective strategies for the active removal or servicing of uncooperative targets have become essential for sustainable space operations \cite{shan2016}.

Active Debris Removal and On-Orbit Servicing missions need to precisely operate near the targets, demanding autonomous systems that can safely perform Rendezvous and Proximity Operations (RPO) \cite{fehse2003}. These operations require sophisticated algorithms capable of identifying, approaching, and maintaining stable relative positions with uncooperative targets \cite{psomopoulos2020}. Several past space missions demonstrated technological capabilities for RPO around known cooperative and uncooperative targets since 1997. Recently, in 2025, the Mission Extension Vehicle (MEV-1), designed by Northrop Grumman, undocked from IS-901 by pushing it into a graveyard orbit in GEO after keeping it in service for almost five years, since its docking in 2020 \cite{space2025satellites}. Similarly, in 2024, the ADRAS-J mission by Astroscale and JAXA demonstrated its RPO capabilities by safely performing three fixed-point observations and fly-around maneuvers around a uncooperative upper stage rocket body \cite{astroscale2024adras}.

These past missions have demonstrated the capabilities of RPO missions with a known object. The problem arises with the uncooperative nature of the targets, because the shape, size, and motion characteristics are unknown \cite{mahendrakar2023spaceyolo} \cite{mahendrakar2024unknown} \cite{floresabad2014}. The complexity increases exponentially when dealing with debris with unpredictable rotation rates. This requires a high-fidelity model of the object to correctly identify the shape and size of the object before docking or de-orbiting the object can be considered \cite{seube2019}.

Many works in computer science have challenged the idea of creating high-fidelity models of a object using only imagery \cite{aharchi2020review,Fan_2017_CVPR,lei2020pix2surflearningparametric3d,moons20103d,stier2023finerecondepthawarefeedforwardnetwork}. The reconstruction of shape from imagery has been a long topic of discussion in various research fields. However, few works have done reconstruction in space environments where lighting conditions pose the biggest obstacle in reconstruction tasks. For effective reconstruction from imagery the lighting conditions and the images need to be detailed and complete, in order for most 3D reconstruction algorithms to operate efficiently for those tasks. Previous works have used 3D reconstruction algorithms for space environments \cite{huber2024high}, but they assume a static scene with controlled lighting. 

This research builds upon the previous works by relaxing the constraint of static models for reconstruction opting for a dynamic more realistic approach utilizing dynamic 3D reconstruction neural networks adapted for this task in space scenarios. This work aims to create high-fidelity models of space objects for the use of bolstering space situational awareness in low earth orbits.

\section{Related Work and Contributions}

 Neural Radiance Field (NeRF) \cite{mildenhall2021nerf} and 3D Gaussian Splatting (3DGS) \cite{kerbl20233d} emerged as promising solutions for reconstructing a 3D scene from 2D images. While, both techniques excel at novel view synthesis, they present significant limitations for space applications. NeRF assumes static scenes and demands substantial computational resources\cite{mergy2021vision}. On the other hand, 3DGS can render in real-time (>=30 fps) but relies on the Structure from Motion (SfM) for camera poses and sparse point cloud \cite{issitt2025optimal}. 
 
 Recent studies have adapted NeRF and its variants to synthesize novel views of uncooperative space targets. Accelerated NeRFs like Instant-NGP and 3DGS were able to render high-quality novel views of the RSO's within few minutes even on edge computing devices with limited computing resources \cite{caruso20233d}. Another study employed prior depth information to render unseen viewpoints using NeRF and then used unseen viewpoints along with seen viewpoints to improve the geometric accuracy of the reconstruction from limited viewpoints \cite{depth_fu2024neural}. 
 
 While these above mentioned methods need camera poses or sparse point clouds, another line of work explores the use of Convolutional Neural Networks for extracting a coarse 3D shape of the RSO, while estimating the pose from a single image using superquadrics \cite{park2024rapid}. The extracted 3D shape reconstruction quality was improved in \cite{bates2025removing} by addressing the body axis ambiguities either through a loss function or removing it, by defining body axis in a canonical form. The above extracted coarse shape was further utilized as a prior for initializing the gaussian in 3DGS \cite{hucfast}.
 
 3D reconstruction of a tumbling uncooperative target in space environment requires robust methods to handle complex motion, extreme lighting conditions, and sensor noise. Existing methods evaluate the application of D-NeRF, 4DGS, which are variants of NeRF and 3DGS that can handle dynamic scenes \cite{nguyen2024characterizing}. They consider a tumbling mock satellite spinning around one axis, but with a stationary camera, simulating a station keeping. Such configurations could be considered as static scene similar to the fly-around maneuver. The scenario where both RSO is tumbling and observer spacecraft equipped with a monocular camera performing fly-around maneuver, which constitutes of a dynamic scene with dual motion, remains significantly underexplored and presents unique challenges for reconstruction. To our knowledge, this would be the first work that explores above mentioned scenario.

 To this extent we consider 4 state-of-the-art NeRF variants for our evaluation. Dynamic NeRF (D-NeRF) \cite{pumarola2021d}, introduces time as an extra parameter to the system. It initially encodes a canonical representation of the static scene and then learns a deformation field that maps this canonical space into the dynamic scene at any given point of time. Similarly Nerfies \cite{park2021nerfies}, also uses a deformation field that wraps the dynamic scene back to its canonical form. Nerfies assigns a learnable code for each image so that the deformation network learns the transformation which is specific to that image, and maps the scene to match its canonical form. It also employs a coarse-to-fine optimization strategy during training, that prevents the optimization from getting stuck by starting with smooth deformations before dealing with detailed deformations. They also use an elastic regularization to ensure physically plausible deformations. While these methods handle deformations efficiently they increase the computational time as they represent deformation fields using MLP. 
 
 To accelerate the training, we choose two additional state-of-the-art NeRF variants. TiNeuVox\cite{fang2022fast}, introduces a hybrid approach that combines time-aware voxel features with a compact deformation network and a radiance network, along with a multi-distance interpolation method that could handle varying motion scales. HexPlane \cite{cao2023hexplane} takes a different approach from the aforementioned methods. Instead of representing a dynamic scene using deep neural networks or voxel based representations, it represents a dynamic 3D scene using six 2D planes and then fuses learned feature vectors from these six planes. The combined feature vector is then passed to a small MLP based radiance network. These approaches reduces the computational costs significantly without compromising the rendering quality.

 These algorithms can be broadly classified into deformation based (D-NeRF, and Nerfies), hybrid-voxel based (TiNeuVox), and hybrid-planes based (HexPlane). By evaluating each state-of-the-art methods from different architectural category, our study provides a comprehensive understanding of how the NeRF variants for dynamic scenes handles the complex challenges of dynamic scene reconstruction of uncooperative RSOs.

\section{Methodology}
 In this paper, we aim to assess the geometric accuracy of the 3D reconstruction of a static scene by comparing the extracted mesh of the reconstructed model with its 3D CAD model. We also evaluate above mentioned state-of-the-art methods for dynamic scene reconstruction of a tumbling uncooperative RSOs, observed from a static position (station keeping). In addition, we investigate the performance of these methods in the challenging scenario where the target exhibits tumbling motion and the observing spacecraft performs fly-around maneuver, addressing the existing gap in the literature.

\subsection{Simulation Framework}
 Leveraging NVIDIA's Isaac Sim, our simulation framework generates physically accurate simulations of the RSO and observer spacecraft maneuvers for both static and dynamic scenes, providing high-fidelity ground truth data which is essential for our study. While Isaac Sim is primarily developed for robotic simulations, we have adapted its capabilities to the space domain. The platform's built-in synthetic data generator offers several distinct advantages over existing simulators; including physics-based rendering with realistic lighting conditions, automatic generation of ground truth data which includes depth maps, semantic segmentation, pose information, and bounding boxes. 

 Our simulator consists of RSO referred as chief at the origin of Hill frame and an observer spacecraft referred as deputy performing fly-around maneuvers at a relative distance. Distant light is used to simulate the sun like lighting conditions. 3D CAD models of the chief are obtained from the NASA resources and turbosquid websites \cite{nasa}\cite{nasa2}\cite{turbosquid}. Specifically, we have considered three satellites for our simulation framework, a 1RU Generic CubeSat, Deformation, Ecosytem Structure and Dynamics of Ice (DESDynI), and Geostationary Operational Environmental Satellites (GOES - R). 
 
 For creating the dynamic dataset, we used a constant rotation rate of 3 deg/s along z-axis as a starting point. We also plan to test these algorithms with constant rotation rates upto 10 deg/s and along multiple axis, to study if these algorithms could handle complex tumbling scenarios.

\begin{figure}[H]
    \centering
    \includegraphics[width=0.3\textwidth]{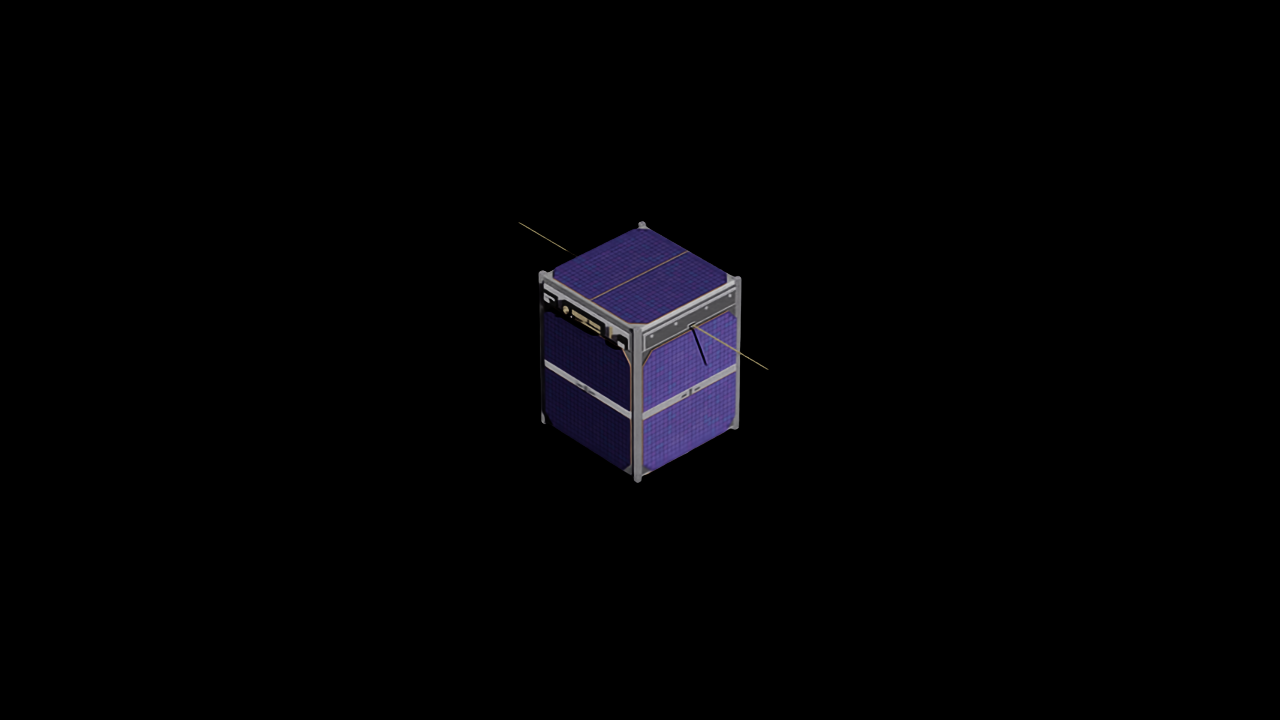}
    \includegraphics[width=0.3\textwidth]{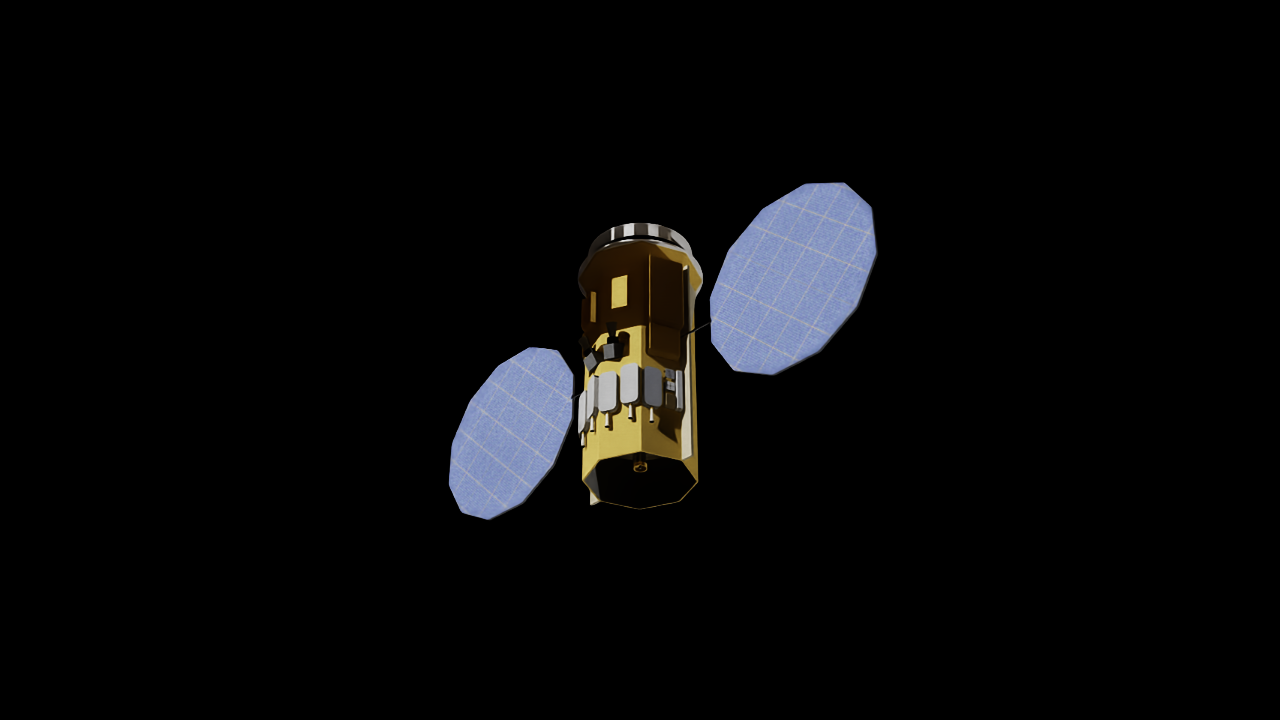}
    \includegraphics[width=0.3\textwidth]{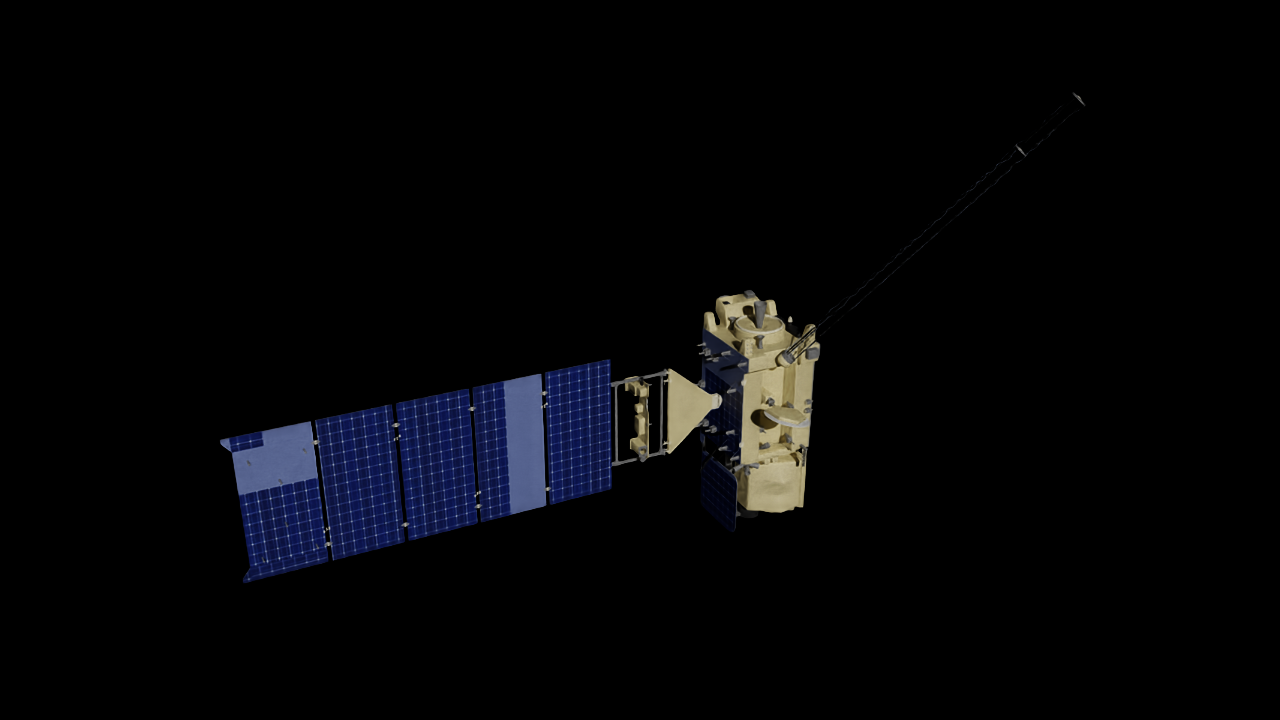}
    \caption{Rendered images of the 3D CAD models in Isaac sim}
    \label{fig:three_images}
\end{figure}

 The details of relative orbits has been presented in the following section.

\subsection{Relative Orbit Dynamics}

Since a major component of a successful reconstruction is knowing the pose of the RSO relative to the camera, and that ideally the entirety of the RSO is imaged, we leverage the dynamics of the system in order to aid this. The RSO being imaged is assumed to be in a circular LEO orbit and that the satellite performing the imaging is close to the RSO, relative to the orbital radius of the RSO. Hence all the assumptions to use the Clohessy Wiltshire (CW) relative equations are satisfied. In order to define the equations of motion, the Inertial frame, $\mathcal{N}$ is assumed to the be Earth Centered Inertial (ECI) frame, whilst the relative Hill frame $\mathcal{O}$, \cite{Schaub2014-op} centered at the RSO of interest and oriented such that $\hat{\mathbf{o}}_r$ is the orbital radius direction, $\hat{\mathbf{o}}_h$ is parallel to the orbital angular momentum of the RSO, and $\hat{\mathbf{o}}_\theta$ completes the right-handed coordinate system. Mathematically, the Hill frame is described as:

\begin{subequations}
    \begin{equation}
        \hat{\mathbf{o}}_r=\frac{\mathbf{r_c}}{r_c}
    \end{equation}
    \begin{equation}
        \hat{\mathbf{o}}_\theta=\hat{\mathbf{o}}_r\times\hat{\mathbf{o}}_h
    \end{equation}
    \begin{equation}
        \hat{\mathbf{o}}_h=\frac{\mathbf{h}}{h}
    \end{equation}
\end{subequations}

Where $\mathbf{r_c}$ is the radial position of the RSO and $\mathbf{h}$ is its angular momentum vector.

The CW equations, which represent the linearized relative equations of motion can then be expressed in continuous-time state-space form as follows:

\begin{equation}
    \boldsymbol{\dot{x}}=
    \begin{bmatrix}
        \dot{x} \\ \dot{y} \\ \dot{z} \\ \ddot{x} \\ \ddot{y} \\ \ddot{z}
    \end{bmatrix} = \boldsymbol{Ax + Bu} = 
    \begin{bmatrix}
        0 & 0 & 1 & 0 & 0 \\
        0 & 0 & 0 & 1 & 0 \\
        0 & 0 & 0 & 0 & 1 \\
        3\dot{f}^2 & 0 & 0 & 2\dot{f} & 0 \\
        0 & 0 & -2\dot{f} & 0 & 0 \\
        0 & 0 & \dot{f}^2 & 0 & 0
    \end{bmatrix}
    \begin{bmatrix}
        x \\ y \\ z \\ \dot{x} \\ \dot{y} \\ \dot{z}
    \end{bmatrix} + 
    \begin{bmatrix}
        0 & 0 & 0 \\ 0 & 0 & 0 \\ 1 & 0 & 0 \\ 0 & 1 & 0 \\ 0 & 0 & 1  
    \end{bmatrix}
    \begin{bmatrix}
        u_x \\ u_y \\u_z
    \end{bmatrix}
\end{equation}

Where $x, y, z$ are in the $r, \theta, h$ directions respectively. 

Until now, we have been using uncontrolled orbits that stably orbit the RSO because of selectively chosen initial conditions resulting in $\mathbf{u}=\mathbf{0}$. In order to generate an orbit that is able to capture the greatest proportion of the RSO in these unstable orbit, the orbit is required to be inclined at 45 degrees to the vertical plane of the RSO. It can be best seen with the CubeSat dataset, where if the relative orbit of the observer spacecraft were planar with RSO, then the top and bottom surfaces of the satellite would never be imaged. By inclining the relative orbit to 45 degrees, this simple trajectory enables us to capture the most of the object. 

The initial conditions for a stable uncontrolled relative orbit as given in \cite{Schaub2004-cd} must satisfy the condition,

\begin{equation}
    \dot{y}_0 = 2\dot{f}x_0
\end{equation}

which gives rise to the initial $\dot{y}_0$ value given some initial $x_0$ value. All other states are set to zero for this bounded orbit. In order to create the inclined orbit, the initial condition can be premultiplied by a direction cosine matrix (DCM), $R$, explaining a rotation of $90^\circ$ around the $\mathbf{o}_h$ and a $45^\circ$ rotation around $\mathbf{o}_r$, as:

\begin{equation}
    R_{\theta,\phi,\psi} = R(\theta)R(\phi)R(\psi)=\begin{bmatrix}
        c(\theta)c(\phi) & c(\theta)s(\phi)s(\psi) - s(\theta)c(\psi) & c(\theta)s(\phi)c(\psi) + s(\theta)s(\phi) \\
        s(\theta)c(\phi) & s(\theta)s(\phi)s(\psi) + c(\theta)c(\psi) & s(\theta)s(\phi)c(\psi) - c(\theta)s(\phi) \\
        -s(\phi) & c(\phi)s(\psi) & c(\phi)c(\psi)
    \end{bmatrix}
\end{equation}

Where $c(), s()$ represent the trigonometric functions $sin()$ and $cos()$ respectively and $\theta=45^\circ$, $\phi=0^\circ$, $\psi=90^\circ$. We can now describe the initial conditions for the inclined, stably bounded relative orbit at some distance $x_0$ from the RSO, as:

\begin{equation}
    \mathbf{x}_0 = R_{90,0,45}\begin{bmatrix}
        x_0 \\ 0 \\ 0 \\ 0 \\ -2\dot{f}x_0 \\ 0
    \end{bmatrix} = \begin{bmatrix}
        0 \\ x_0 \\ 0 \\ \frac{2\dot{f}x_0}{\sqrt{2}} \\ 0 \\ \frac{2\dot{f}x_0}{\sqrt{2}}
    \end{bmatrix}
\end{equation}

\begin{figure}[h]
    \centering
    \includegraphics[width=0.5\linewidth]{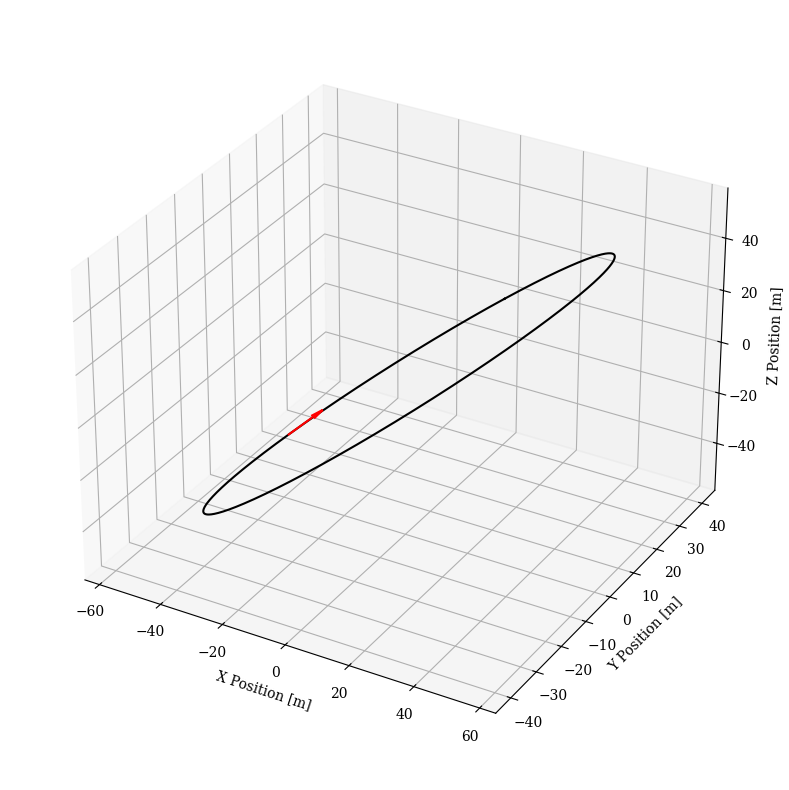}
    \caption{45$^\circ$ Inclined Orbit of Relative Motion between Observer Spacecraft and RSO}
    \label{fig:trajectory}
\end{figure}

The evolution of the relative dynamics under these initial conditions although stable and bounded, produce a trajectory that is elliptical with a major axis twice that of the minor axis as seen in Figure \ref{fig:trajectory}. No such initial conditions create a stably bounded circular orbit, which would be ideal for imaging conditions. Having an elliptical orbit means that as the satellite moves further from the RSO, less of the RSO occupies the image field which could in turn degrade quality in reconstruction or create issues with focal lengths that also degrade the reconstruction quality. As such, we will continue to investigate a means for creating circular relative orbits to improve the reconstruction quality. It is noted, that for the purposes of the current simulation, an initial distance from the RSO is chosen as $x_0=40m$.

\subsection{Evaluation Metrics}

To evaluate reconstructed dynamic scene we will use Peak Signal to Noise Ratio (PSNR) and Structural Similarity Index Measure(SSIM) and Learned Perceptual Image Patch Similarity (LPIPS). 
PSNR is a measures how closely rendered image matches the ground truth image at a pixel level. While SSIM evaluates the perceived change in structural information between  the rendered image and ground truth image by considering luminance and contrast, rather than just measuring pixel-to-pixel differences. LPIPS uses pre-trained CNNs on image claasification tasks to measure perceptual differences between the images. Higher PSNR and SSIM values indicates better quality reconstruction, while lower LPIPS indicates that the rendered images are more perceptually similar to the ground truth.

\section{Preliminary Results}

\begin{figure}[h!]
\centering
  \includegraphics[width=.4\linewidth]{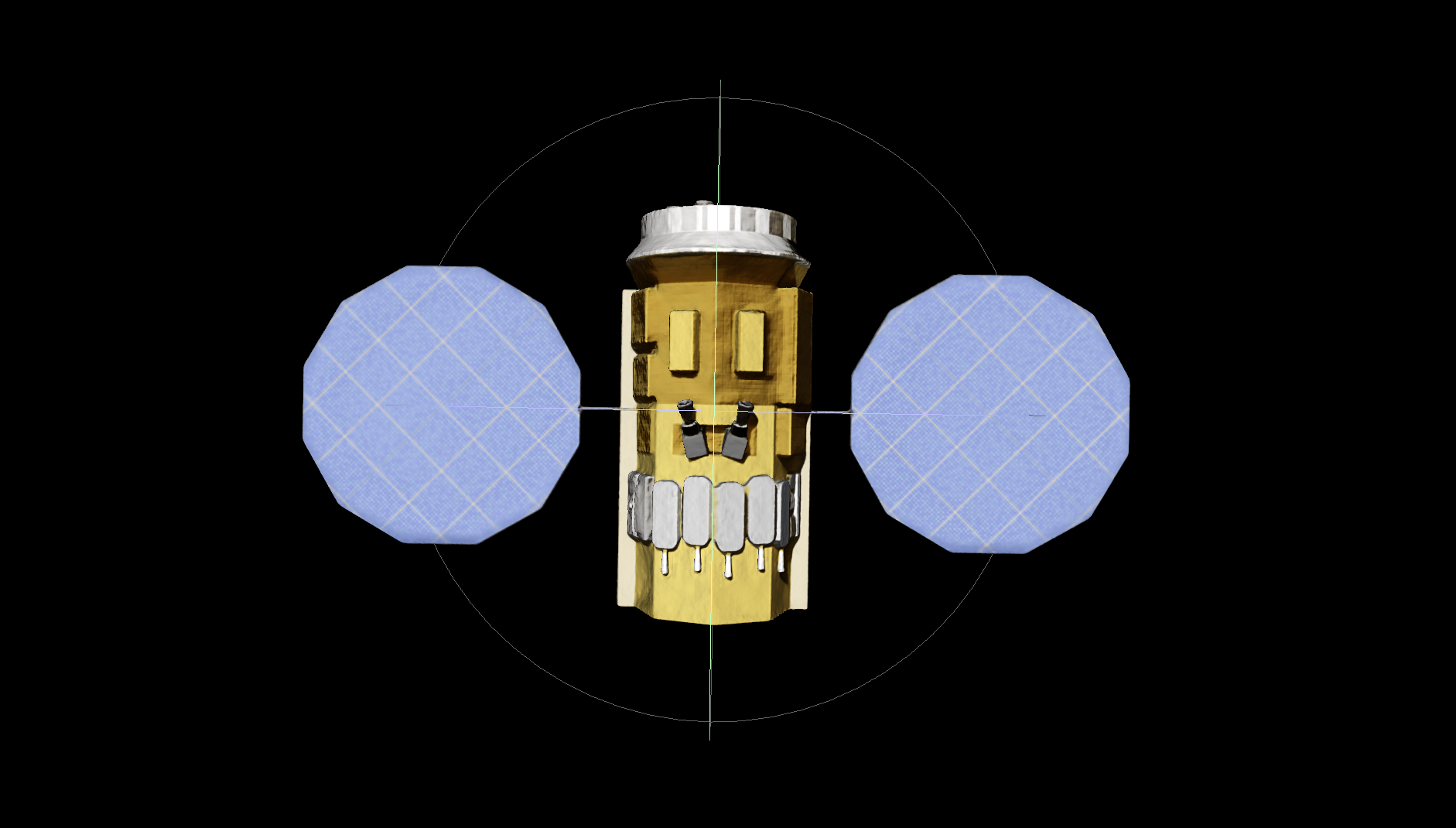}
  \label{fig:cylinder40mResult1_mesh}
  \centering
  \includegraphics[width=.4\linewidth]{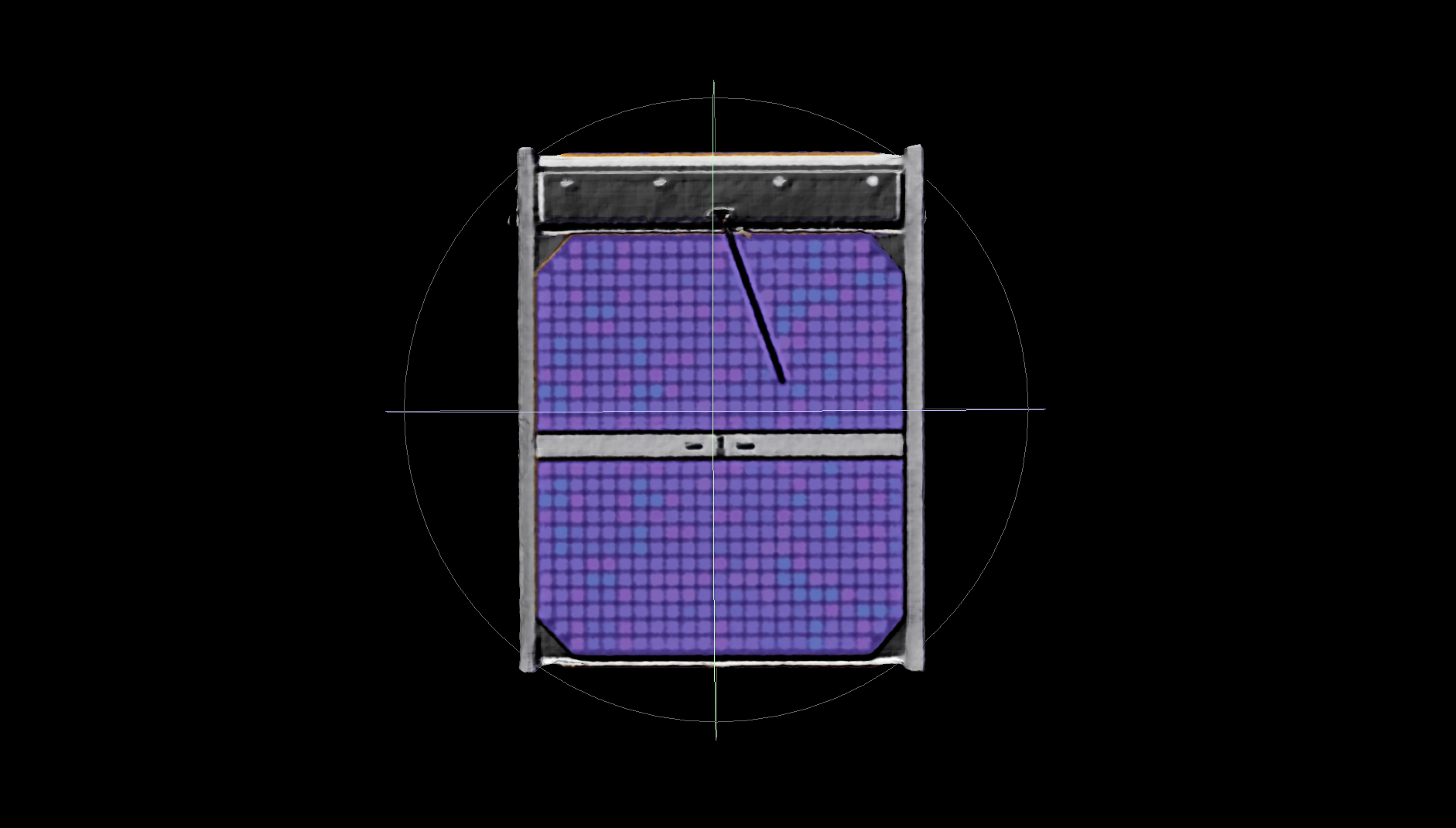}
  \label{fig:cubesat40mCCresult_mesh}
\caption{Reconstructed mesh for DESDynl and Cubesat, with textures}
\label{fig:MeshResults}
\end{figure}

\begin{figure}[h!]
\centering
\includegraphics[width=0.4\linewidth]{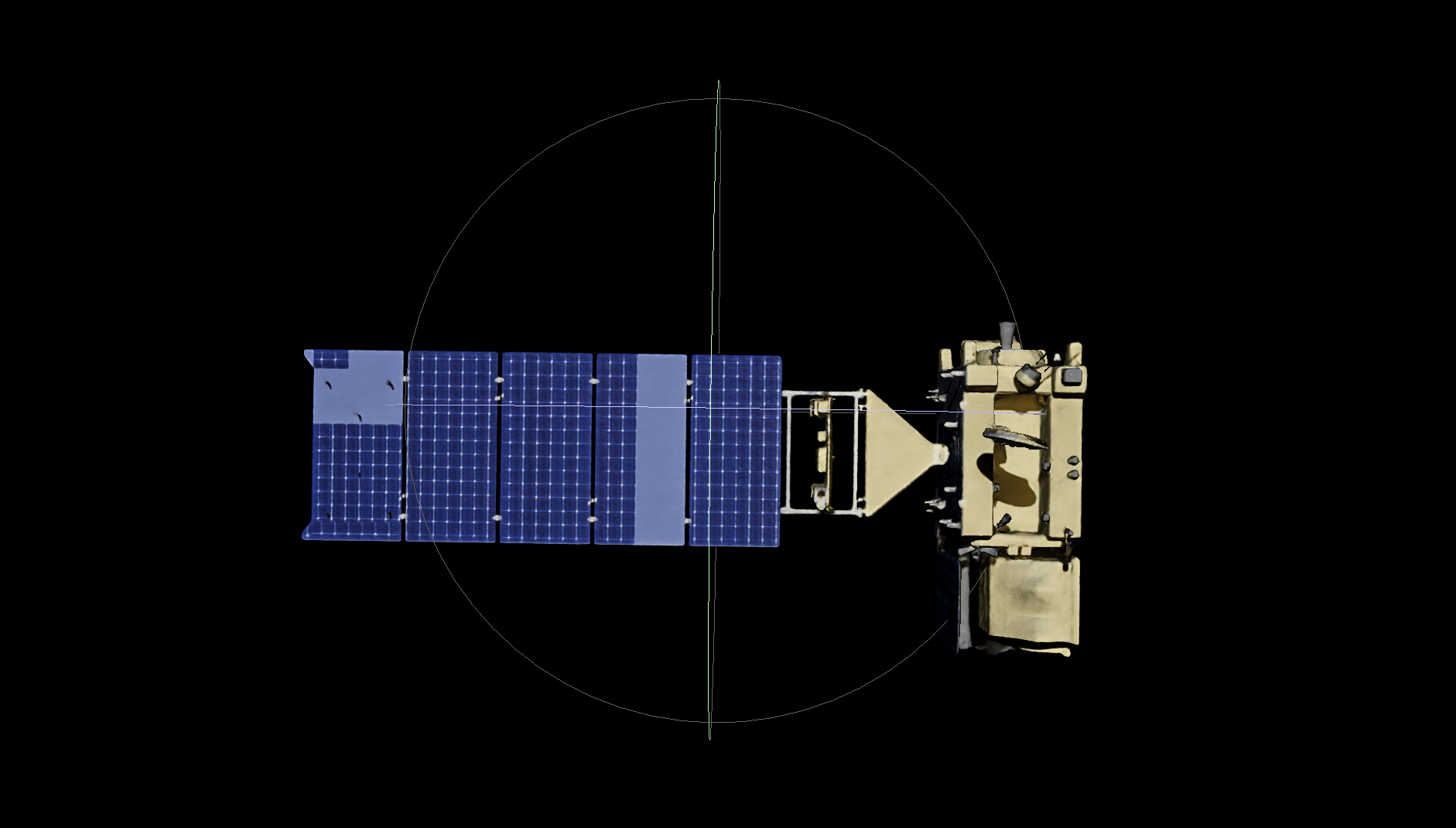}
\caption{\label{fig:GOES40mResult1_mesh}Reconstructed mesh for GOES-R satellite, with textures}
\end{figure}

We initially wanted to evaluate the geometric accuracy for a static scene, a scenario where RSO is not spinning, so we generated three synthetic datasets one for each considered 3D CAD model. We tested these datasets using Neuralangelo \cite{li2023neuralangelo}, a 3D reconstruction algorithm developed by Nvidia, its strengths lies in applying coarse-to-fine optimization to multi-hash encoding introduced in instant-ngp. Through this architecture neuralangelo is capable of creating high fidelity surface meshes from 2D images, this can be seen in Figures \ref{fig:MeshResults} \ref{fig:GOES40mResult1_mesh}. We used MeshLab \cite{cignoni2008meshlab} to visualize generated meshes. The geometric accuracy was measured using cloud to mesh distance metric in cloud compare software \cite{cloudcompare2021}, the results are presented below.

\subsection{Static Scene 3D Reconstruction using Neuralangelo}
Results analysis will take the form of both objective comparisons between generated reconstructions and reference models as well as subjective observations on general trends in the performance on certain features or under certain conditions. Objective comparisons will be made by calculating Euclidean distances from point clouds sampled from generated reconstruction meshes to the reference models used. These distances will be referred to as Cloud to Mesh, or C2M, distances. Signed distances are used to differentiate between parts of the reconstruction lying within the reference model and parts of the reconstruction lying outside of the reference model

From early tests using the simulation environment already built, preliminary results have been obtained. In the future, these individual results can be normalized by model scale and cumulative results can be analyzed instead of looking at individual examples. Example heatmaps of C2M signed distance scalar fields for the generated reconstructions can be seen in Figure~\ref{fig:C2MResults}. On a broader note, the results we have seen so far are quite promising and could be useful at some level for characterization of the RSO. One of the more common issues seen is with the reconstruction of thin objects, which commonly don't show up at all in reconstructions. This can pose an issue for capturing certain instruments on spacecraft.

\begin{figure}[h!]
\centering
  \includegraphics[width=.4\linewidth]{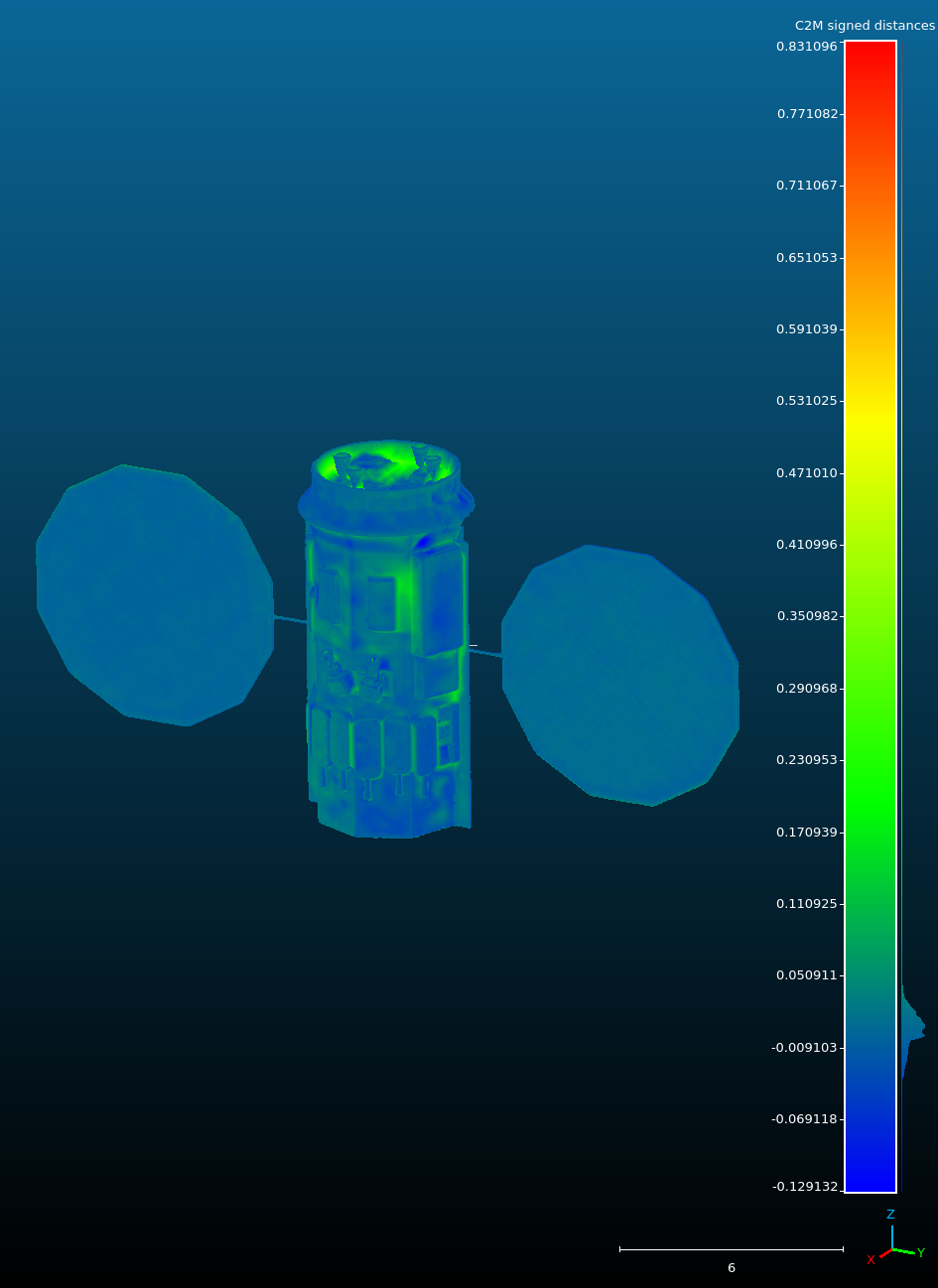}
  \label{fig:cylinder40mResult1}
  \centering
  \includegraphics[width=.4\linewidth]{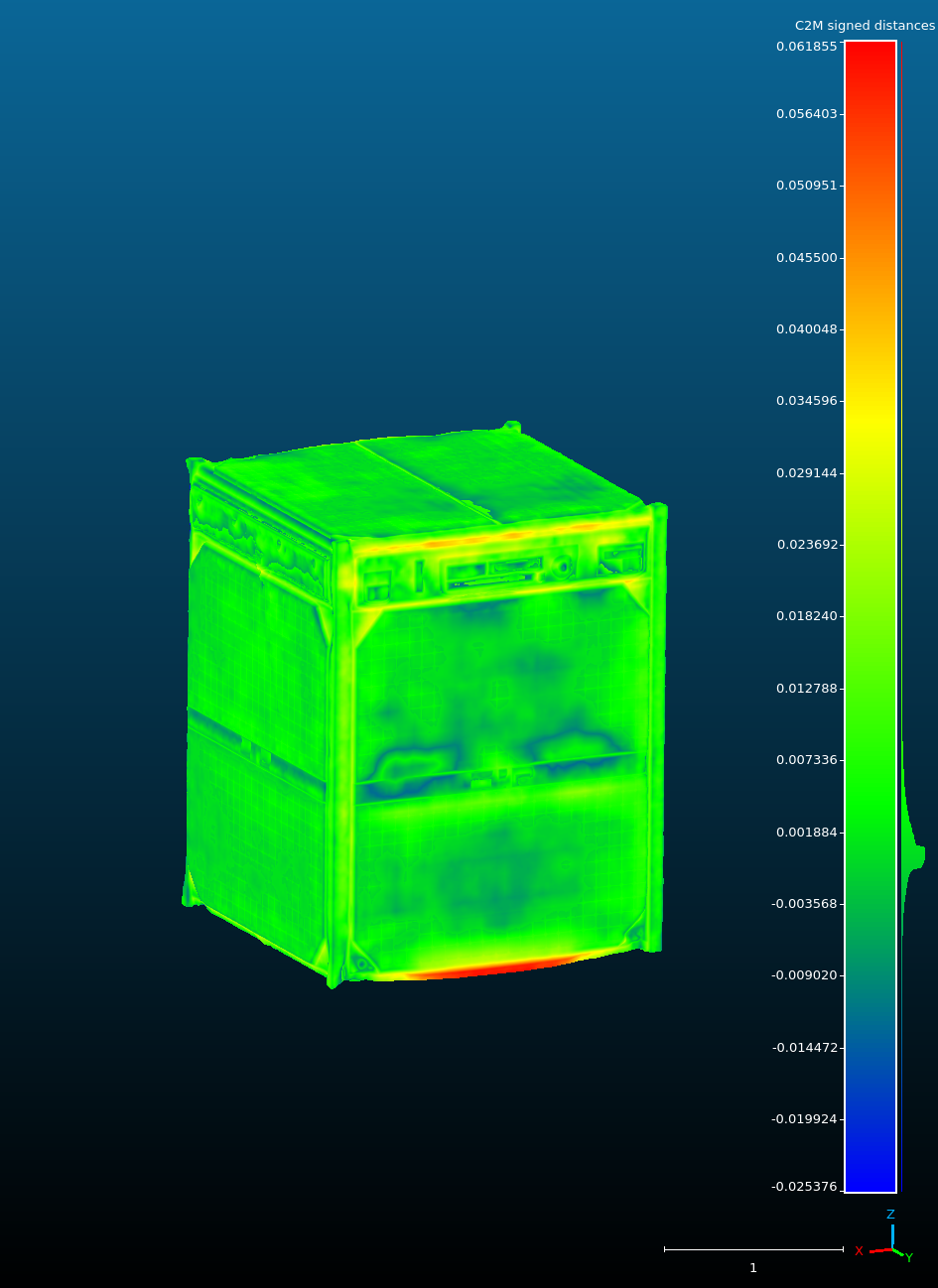}
  \label{fig:cubesat40mCCresult}
\caption{Heatmap of C2M signed distances on the surfaces of reconstructed meshes.}
\label{fig:C2MResults}
\end{figure}

\begin{figure}[h!]
\centering
\includegraphics[width=0.4\linewidth]{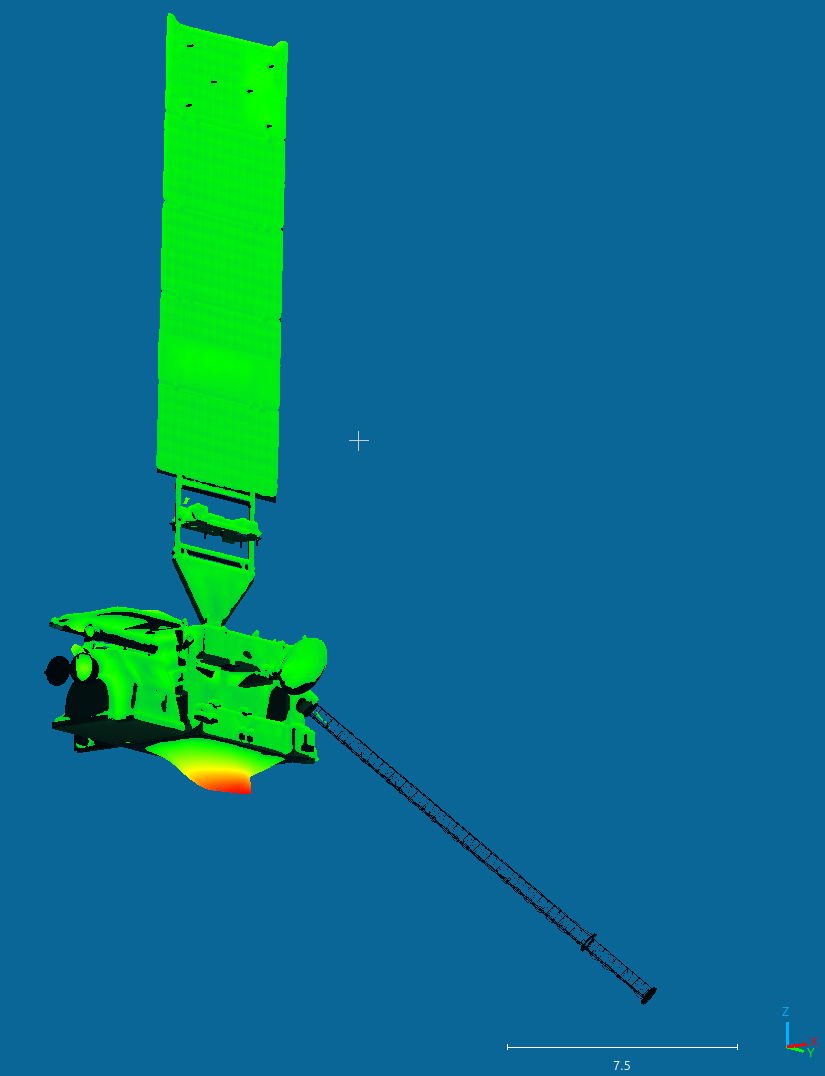}
\caption{\label{fig:GOES40mResult1}Example of the shortcoming of the current application of C2M distance for analysis. The reference model is in black and the C2M signed distance scalar field for the reconstructed mesh is represented by a heatmap.}
\end{figure}

One potential issue associated with the analysis method currently employed is the lack of recognition of errors which occur when a reconstruction simply lacks a part of the original model. An example of this issue can be seen in Figure~\ref{fig:GOES40mResult1}, where a magnetometer on a GOES-R satellite, seen in the bottom right of the figure, is completely missing from the reconstruction. This error isn't represented in the calculated scalar field for the reconstructed mesh as the closest available point on the reference mesh was on the bus of the spacecraft. One analysis method we may switch to in order to eliminate such oversights in the future is calculating the distance from the reference mesh to the reconstruction mesh, not the other way around.

For each reconstruction, C2M signed distances can be displayed as a histogram, allowing for an analysis of the distributions of errors. This can be helpful to compare tails in the error distributions, which is important as tails in the error distribution tend to represent an outsized issue in reconstructions from a subjective standpoint. An example of such a histogram, with a Gaussian distribution fitted to the data, can be seen in Figure~\ref{fig:C2MResults}. As normal distributions don't seem to fit the data too well for most experiments done so far, in the future we may switch our analyses to use a different distribution, such as Weibull distributions. Moreover, neuralangelo took around 8 hours to produce these high fidelity meshes with textures.

\begin{figure}[h!]
\centering
\includegraphics[width=0.6\linewidth]{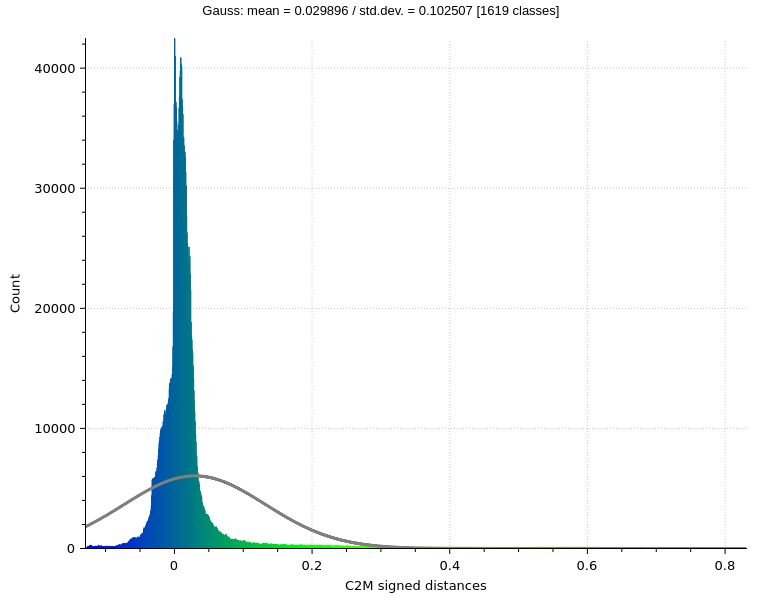}
\caption{\label{fig:C2MResults}Histogram of C2M signed distances for a reconstructed mesh of DESDynl 3D CAD model.}
\end{figure}

\pagebreak
\section{Expected Results \& Conclusion}

In this paper, we aim to evaluate the performance of four state-of-the-art 3D reconstruction algorithms for a tumbling RSO while the observer spacecraft performs a fly-around maneuver. This is done by simulating the relative dynamics between the RSO and the observer spacecraft under realistic space conditions in Isaac Sim. This simulation framework would allow us to generate synthetic datasets for this assessment, across multiple scenarios, with different camera parameter, observation distances, and rotation rates. We expect that our analysis would provide key insights into the performance trade-offs between the computational efficiency and reconstruction fidelity of these algorithms, with the goal of rendering a scene representation that preserves the geometry of the RSO. Furthermore, we plan to investigate the optimal trajectories which would gives us optimal viewing points that could potentiality improve the reconstruction quality. Our initial findings provide a strong baseline for our current work in characterization of RSOs using vision sensors like monocular camera with applications in rendezvous proximity operations and docking, on-orbit servicing, and active debris removal. The results and findings results could potentiality establish foundations for specialized algorithms that could handle challenges related to space-based 3D reconstruction.

\nocite{*}
\bibliography{references}

\end{document}